\documentclass[10 pt, conference]{ieeeconf}      % Use this line for a4

\let\proof\relax \let\endproof\relax                                    

\usepackage[utf8]{inputenc} % on veut nos accents !

\IEEEoverridecommandlockouts                              % This command is only
\usepackage{graphics} % for pdf, bitmapped graphics files
\usepackage{epsfig} % for postscript graphics files
\usepackage{times} % assumes new font selection scheme installed
\usepackage{amsmath} % assumes amsmath package installed
\usepackage{amssymb}  % assumes amsmath package installed
\usepackage{amsthm}
\usepackage{color}
\usepackage[hidelinks]{hyperref}

\usepackage[Algorithm]{algorithm}
\usepackage{algpseudocode}

\graphicspath{{./figures/}}
\DeclareGraphicsExtensions{.pdf,.png,.eps,.ps}

\newtheorem{theorem}{Theorem}

\newtheorem{property}{Property}

\newcommand{\otn}{{\{1...n\}}}

\newcommand{\chifree}{\chi^{\mathrm{free}}}
\newcommand{\chiobs}{\chi^{\mathrm{obs}}}
\newcommand{\chigoal}{\chi^{\mathrm{goal}}}
\newcommand{\chireach}{\chi^{\mathrm{reach}}}

\newcommand{\xdx}{(x\ x'\cdots x^{(p)})}

\renewcommand{\max}{\mathrm{max}}
\renewcommand{\min}{\mathrm{min}}

\newcommand{\dt}{\Delta T}

\newcommand{\init}{\mathrm{init}}

\newcommand{\RR}{\mathbb{R}}

\title{\LARGE \bf
Priority-based intersection management\\
with kinodynamic constraints
}

\author{
Jean Grégoire$^\star$\thanks{$\star$ MINES ParisTech, Center for Robotics, 60 Bd St Michel 75272 Paris Cedex 06, France}\hspace{10mm}
Silvère Bonnabel$^\star$\hspace{10mm}
Arnaud de La Fortelle$^{\star\dagger}$\thanks{$\dagger$ Inria Paris - Rocquencourt, RITS team,  Domaine de Voluceau - Rocquencourt, B.P. 105 - 78153 Le Chesnay, France}
}
\include{myLatex}

\graphicspath{{figures/}}
\begin{document}
\maketitle
\thispagestyle{empty}
\pagestyle{empty}

%%%%%%%%%%%%%%%%%%%%%%%%%%%%%%%%%%%%%%%%%%%%%%%%%%%%%%%%%%%%%%%%%%%%%%%%%%%%%%%%
\begin{abstract}
We consider the problem of coordinating a collection of robots at an intersection area taking into account dynamical constraints due to actuator limitations. We adopt the coordination space approach, which is standard in multiple robot motion planning. Assuming the priorities between robots are assigned in advance and the existence of a collision-free trajectory respecting those priorities, we propose a provably safe trajectory planner satisfying kinodynamic constraints. The algorithm is shown to run in real time and to return safe (collision-free) trajectories. Simulation results on synthetic data illustrate the benefits of the approach.
\end{abstract}

\section{Introduction}

\subsection{Motivation}

Human error is the sole cause in 57\% of all road accidents and is a contributing factor in over 90\% \cite{Treat1977,NCSA2004}. Moreover, traffic congestion motivates the research to improve intersection traffic flow. Intelligent transportation systems are expected to tackle both safety and efficiency issues in the near future. Many systems have been proposed and they have proved their ability to increase traffic efficiency -- particularly compared to traffic light systems -- and to reduce the risk of road accidents \cite{Dresner2004,Dresner2008,Colombo2012,I.Zohdy2012,Kowshik2011,Mehani2007}. Furthermore, more generally, automated conflict management opens new perspectives to improve railway \cite{Ismail1999} and air transportation systems \cite{Tomlin1998} efficiency. In transportation systems, safety is usually centralized (e.g. air traffic control, rail management systems) or at least managed locally in a centralized way (e.g. traffic lights). In the future, we anticipate there will be locally full information, e.g. through car-to-car communication being currently standardized. Obviously there will be non-communicating entities, sometimes delays or sensing errors, but our aim is to go from a centralized system in full information down to more reactive schemes, ensuring safety first.

\subsection{Related work}

The standard approach to multi robot motion planning is to decompose the problem into two parts, as initiated in \cite{Kant1986}. As presented in \cite{Gregoire2012}, the first one consists of determining fixed paths along which robots cross the intersection. The second one consists of computing the velocity profile of each robot along its path: this is a well-known problem studied for applications in automated guided vehicles (AGVs) and robot manipulators.

As first introduced in \cite{Leroy1999,Lavalle1996}, the path-velocity decomposition enables to introduce an abstract space: the coordination space. It is a standard approach to robot motion planning \cite{Latombe1991,Lozano-Perez1980}, and the motion planning problem in the real space boils down to finding an optimal trajectory in the coordination space that is collision-free with respect to an obstacle region. The coordination space is a $n$-dimensional space (where $n$ denotes the number of robots in the intersection) and the obstacle-region has a cylindrical structure \cite{LaValle2006}. In \cite{Gregoire2012}, we have revisited the notion of priorities to propose a novel framework for automated intersection management based on priority assignment. It is a very intuitive notion: the priority graph indicates the relative order of robots. Our framework enables to decompose the motion planning problem problem in the coordination space into a combinatorial problem: priority assignment, and a continuous problem: finding an efficient trajectory with assigned priorities. 

The ambition of this framework is to enable more robustness and distribution in future automated intersection management systems. Indeed, existing intersection management systems such as proposed in \cite{I.Zohdy2012,Dresner2004,Mehani2007} plan the complete trajectories of robots through the intersection and ensuring safety requires robots to follow precisely the planned trajectory. By contrast, if priorities only are planned, the priority graph can be conserved even if some unpredictable event requires a robot to slow down for some time.

It is now clear that the combinatorial problem of assigning efficient priorities is inherently difficult, as noticed in \cite{Ghrist2006} and developed in the priority-based framework in \cite{Gregoire2012}. As a result, we will only consider in the present paper the issue of planning "good" trajectories for already assigned priorities. When the robots can start and stop instantaneously, it is relatively easy to define an optimal trajectory for fixed priorities. This trajectory is referred to as the left-greedy trajectory \cite{Ghrist2006,Gregoire2012}. However, taking into account acceleration (and higher derivatives) constraints turns the optimization problem into a "highly non-trivial" problem (as suggested in the conclusion of the paper \cite{Ghrist2006}).  In the present paper, we address the challenging problem of finding safe trajectories that respect this type of constraints. In \cite{Peng2005}, the problem is formulated as a mixed integer nonlinear programming problem, and the solution proposed is suitable only for a "reasonable" and fixed number of robots. Moreover, priority assignment and trajectory planning are not decoupled. In the present paper, we focus on a low complexity solution to the trajectory planning problem with assigned priorities which is applicable for a large and potentially varying number of robots.

\subsection{Contributions}

We introduce a theoretical tool: the braking trajectory, which is a virtual trajectory obtained letting all robots slowing down as much as possible to stop. The key idea of the paper is to ensure that at every time-step, the (virtual) braking trajectory is collision-free. With the proposed planner, robots are maximally aggressive, i.e. always maximize the distance travelled at every time-step. However, they do not accelerate if the virtual braking trajectory becomes unsafe or violates a priority, i.e. they ensure the existence of a failsafe maneuver for the system of robots at any time. We present a trajectory planner with assigned priorities that consists of just-in-time braking and is proved to return collision-free trajectories respecting the assigned priorities.

Section \ref{sec-model} and \ref{sec-priority} present the modelling assumptions and recall the basics of the priority-based framework of \cite{Gregoire2012}. Section \ref{sec-collision-avoidance} introduces the motion planner algorithm along with its safety guarantees. Finally, simulation results of Section \ref{sec-simulations} illustrate the efficiency of the approach.

\section{Modelling assumptions}
\label{sec-model}

\subsection{The coordination space}

We assume that robots are constrained to follow predefined paths to go through the intersection. The paths are not necessarily straight lines: robots are just considered as driving along fixed tracks. This can be achieved by a low-level controller. This standard assumption \cite{Leroy1999,Fraichard1989,Hafner2011,Akella2002,Kowshik2011} fits well intersections in a road network, where robots travel along lanes.

\begin{figure}[ht]
\centering
\includegraphics[width=1\linewidth]{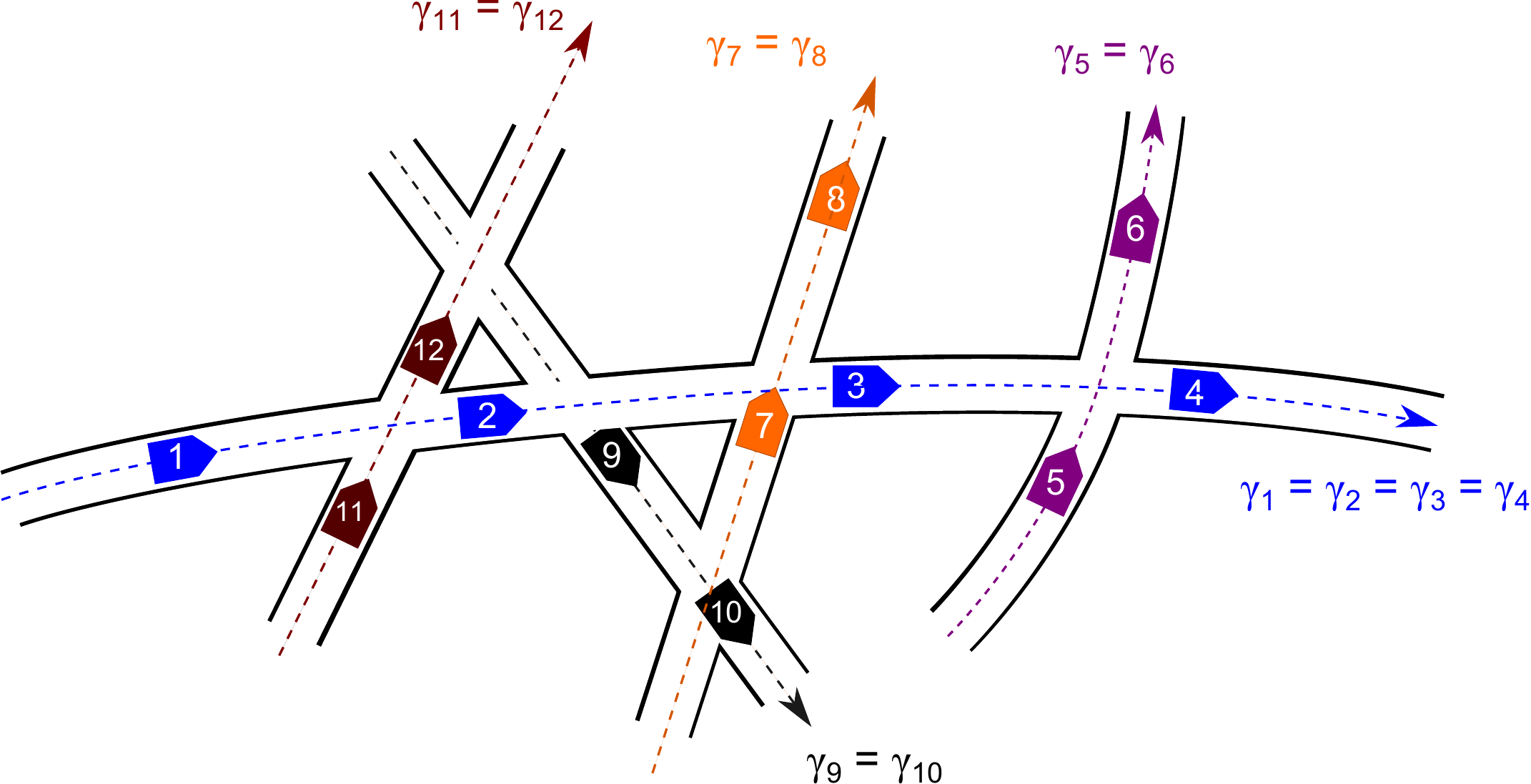}\hfill
\caption{The path following assumption. All robots in the same lane (depicted with the same color) travel along the same geometric path with independent velocity profiles.}
\label{fig-paths}
\end{figure}

Every robot $i$ follows a particular path $\gamma_i$ and we denote $x_i \in \RR$ its curvilinear coordinate along the path. The configuration of the system of robots is ${x=(x_i)_{i\in\otn}}$ and we denote $x(t)$ the evolution of $x$ through time $t\in[0,T]$. Figure \ref{fig-paths} illustrates the path following assumption.

The configuration space $\chi$ is known as the coordination space \cite{ODonnell1989,LaValle2006,Leroy1999}. In the rest of the paper,  $\{\mathbf{e}_i\}_{1\le i\le n}$ denotes the canonical basis of $\chi$. The use of the coordination space and the results of this paragraph are standard \cite{LaValle2006}. As every robot occupies a non-empty geometric region, some configurations must be excluded to avoid collisions between robots. The obstacle region $\chiobs$ is the open set of all collision configurations. $\chifree=\chi\backslash \chiobs$ denotes the obstacle-free space.

\begin{figure}[ht]
\centering
\includegraphics[width=1\linewidth]{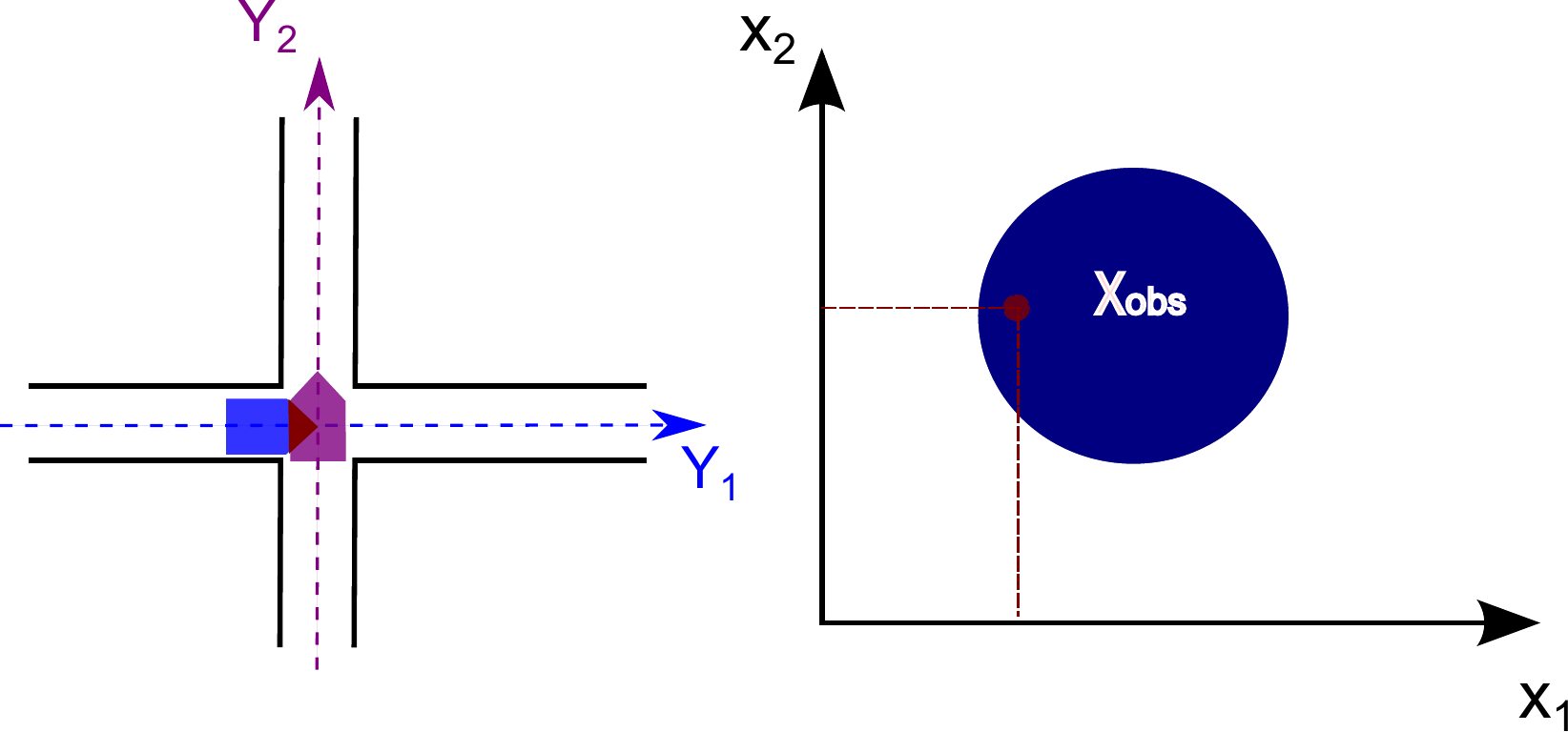}\hfill
\caption{The left drawing depicts two robots in collision in a 2-path-intersection. The right drawing depicts the corresponding configuration in the coordination space that belongs to the obstacle region.}
\label{fig-collision-obstacle-region}
\end{figure}

A collision occurs when two robots occupy a common region of space, so that the obstacle region can be described as the union of $n(n-1)/2$ open cylinders $\chiobs_{ij}$ corresponding to as many collision pairs: $\chiobs = \cup_{i> j} \chiobs_{ij} $. Each cylinder $\chiobs_{ij}$ is assumed to have an open bounded convex cross-section (in the plane generated by $\mathbf{e}_i$ and $\mathbf{e}_j$). Figure \ref{fig-collision-obstacle-region} displays the obstacle region and a collision configuration for a two-path intersection. We assume $\chiobs\neq\emptyset$ (otherwise, coordination is not required), so the boundedness condition ensures that $\inf \chiobs$ and $\sup \chiobs$ both exist.

\subsection{Kinodynamic constraints}

In this paper, we propose to take into account the technical constraints of the robots at the motion planning phase. These include kinematic constraints (maximum velocity, maximum curve radius, etc.) and dynamic constraints (limited acceleration, adherence, jerk, etc.). Let $p$ denote the degree of the constraints and $n$ the number of robots. Let ${s(t)=\xdx(t)}\in\RR^{n\times (p+1)}$ denote the state of the system. We let $x(t)=\pi(s(t))$ denote the first column of the state $s(t)$, that is the position of all robots. We say a trajectory $x$ respects the kinodynamic constraints $C$ if: ${\forall i \in \otn}$, ${\forall t \in [0,T]}$, we have $s_i(t)\in C_i$ with $C_i \subset \RR^{p+1}$ representing the constraints for robot $i$ and $C=\prod_{i\in\otn} C_i \subset\RR^{n\times (p+1)}$.

Note that every robot can have different constraints $C_i$, and $C_i$ can not necessarily be expressed in a product form (for example, the constraint on the velocity can depend on the position).

We assume that the kinodynamic constraints are such that the set of reachable positions from a given state in finite time is bounded. More precisely, the set of reachable positions from state $s_0$ in a time-length $t$:
\begin{equation}
{\chireach(s_0,t)=\left\{x(t)
\left|
\begin{array}{l l}
x \text{ respects the constraints }C \\
s(0)=s_0
\end{array}
\right.
\right\}}
\end{equation} is assumed to be continuous with respect to $s_ 0$ and to be a bounded hypercube of ${x_0+\RR_+^n}$. Note that, the above assumptions imply in particular that:
\begin{enumerate}[]
\item robots cannot travel backwards in the intersection,
\item and from a given state $s_0$, the set of reachable positions in finite time is bounded, and  the bounds depend on the state $s_0$ of the robots (position, velocity, acceleration, etc.).
\end{enumerate}

\section{The priority-based framework}
\label{sec-priority}
In this section, we recall the basics of priority-based intersection management introduced in our previous work \cite{Gregoire2012}.

\subsection{The priority graph}

\begin{figure}[ht]
\centering
\includegraphics[width=1\linewidth]{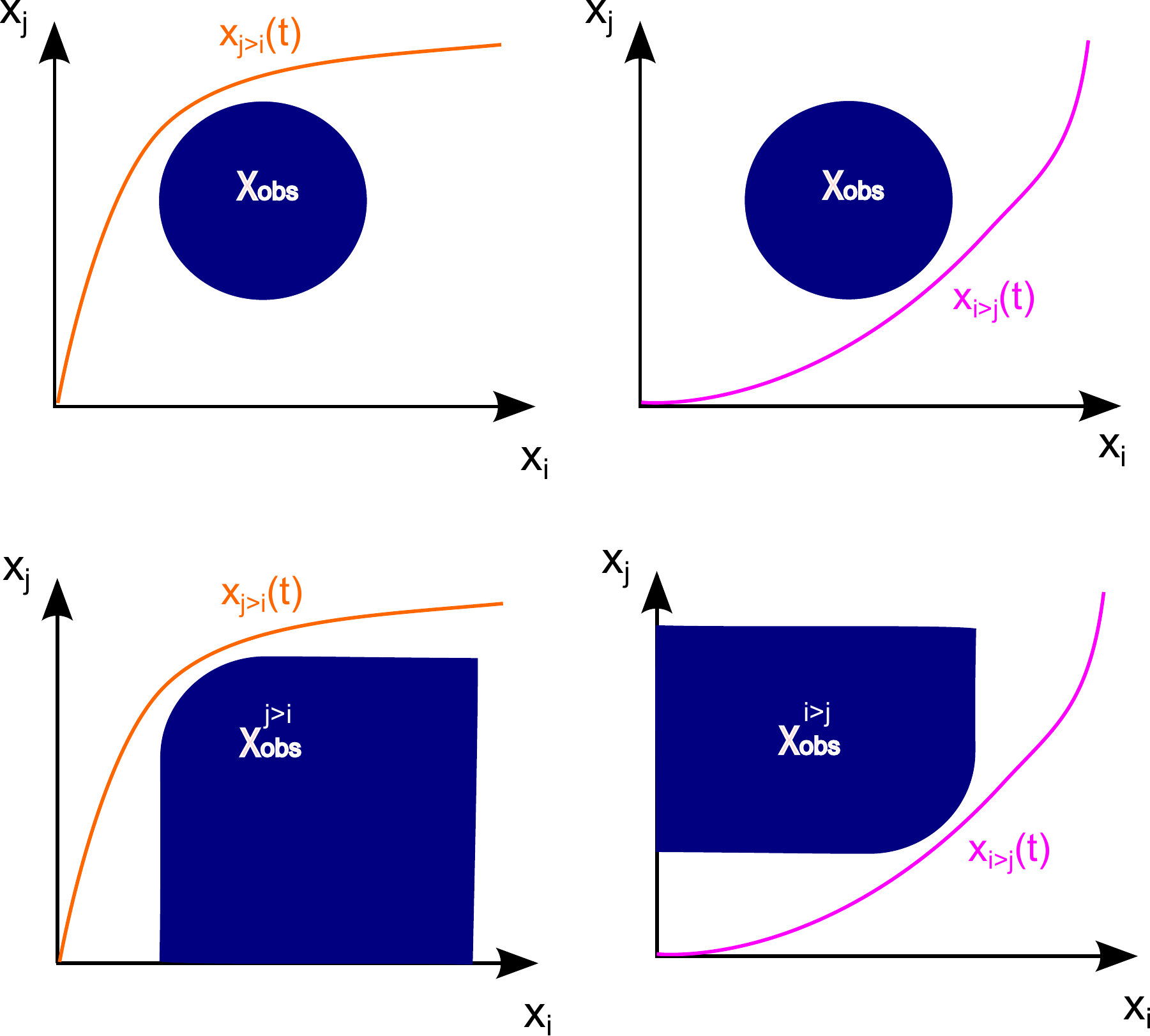}\hfill
\caption{The top drawings represent in the plane $(x_i,x_j)$ the obstacle region $\chiobs$, a collision-free trajectory $x_{i \succ j}$ respecting priority $i\succ j$ and a collision-free trajectory $x_{j \succ i}$ respecting priority $j \succ i$. The bottom drawings depict $\chiobs_{i\succ j}$ and $\chiobs_{j\succ i}$.}
\label{fig-fixed-priorities-collision-cylinder}
\end{figure}

Consider the region $\chiobs_{i\succ j}$ defined as follows and depicted in Figure  \ref{fig-fixed-priorities-collision-cylinder}:
\begin{equation}
\chiobs_{i\succ j} = \chiobs_{ij} - \RR_+ \mathbf{e}_i + \RR_+ \mathbf{e}_j
\end{equation}

We define a natural binary relation corresponding to priority relations between robots. A collision-free trajectory $x$ induces a binary relation $\succ$ on the set $\otn$ as follows. For $i\neq j$ s.t. $\chiobs_{ij}\neq \emptyset$, $i \succ j$ if $x$ is collision-free with $\chiobs_{i\succ j}$.

\begin{figure}[ht]
\centering
\includegraphics[width=.4\linewidth]{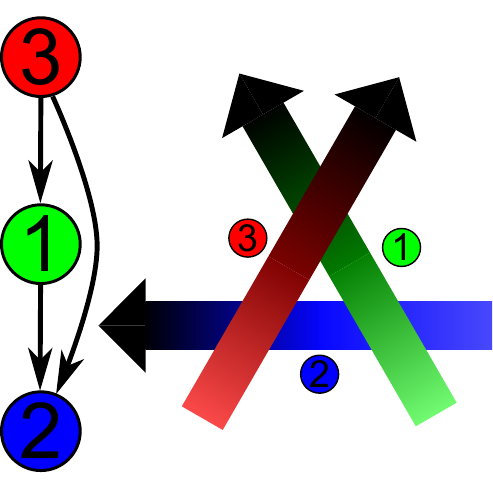}\hfill
\includegraphics[width=.4\linewidth]{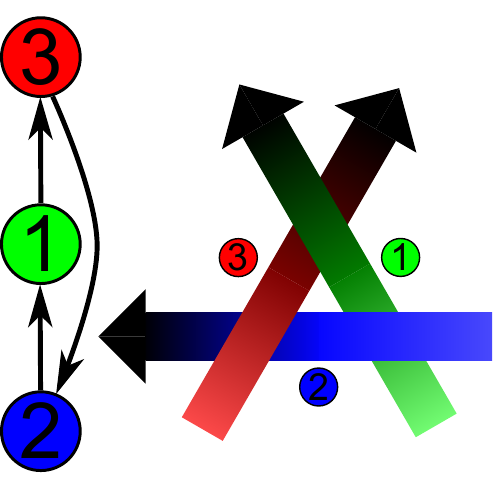}
\caption{Two representations of priority relations. Robots along a path in foreground have priority over robots along a path in background.}
\label{fig-priorities}
\end{figure}

The priority relation can be described by a graph $G$ with nodes $\otn$, where each edge represents the relative priority of a pair of robots. Given a collision-free trajectory $x$, the priority graph is defined as the oriented graph $G$ whose vertices are $V(G)=\otn$ and such that there is an edge from $i$ to $j$ if $i \succ j$, we write $(i,j)\in E(G)$ where $E(G)$ denotes the edge set. An example of a priority graph for 3 robots along 3 distinct paths is described in Figure \ref{fig-priorities}.

\subsection{Problem formulation}

The initial state of the robots is $s^\init$, and the goal region is $\chigoal=\left(\sup \chiobs + \RR_+^n\right) \subset\chifree$. A feasible trajectory for the considered problem is a trajectory $x:[0,T]\rightarrow\chifree$ respecting constraints $C$ such that $s(0)=s^\init$ and $x(T)\in\chigoal$. The multiple robot motion planning problem consists of finding a feasible trajectory. The benefit of the priority-based approach is that the priority graph captures the discrete part of the problem that consists of assigning the relative order of robots through the intersection. 

When the priority graph $G$ is fixed, for all $(i,j)\in E(G)$, the trajectory must be collision-free with regards to $\chiobs_{i\succ j}$. Given, a priority graph $G$, the collision region with regards to priorities $G$ is merely defined as:
\begin{equation}
\chiobs_G = \bigcup_{(i,j)\in E(G)} \chiobs_{i\succ j}
\end{equation}
It is natural to define $\chifree_G=\chi\setminus\chiobs_G$, so that $\{\chifree_G, \chiobs_G\}$ form a partition of $\chi$. In this paper, we focus on the problem on finding a feasible trajectory respecting assigned priorities, i.e. a trajectory $x:[0,T]\rightarrow\chifree_G$ respecting constraints $C$ such that $s(0)=s^\init$ and $x(T)\in\chigoal$.

\section{Motion planner with assigned priorities}
\label{sec-collision-avoidance}
 
The key idea is that if robots wait to be at the boundary of the collision region to brake (as it is the case without dynamic constraints in \cite{Gregoire2012}), collisions will occur because robots can not stop instantly. That is why we need to anticipate the approach of the collision region. This can be done introducing two virtual trajectories as follows.

\subsection{Introducing maximal and minimal trajectories}

\begin{figure}[ht]
\centering
\includegraphics[width=1\linewidth]{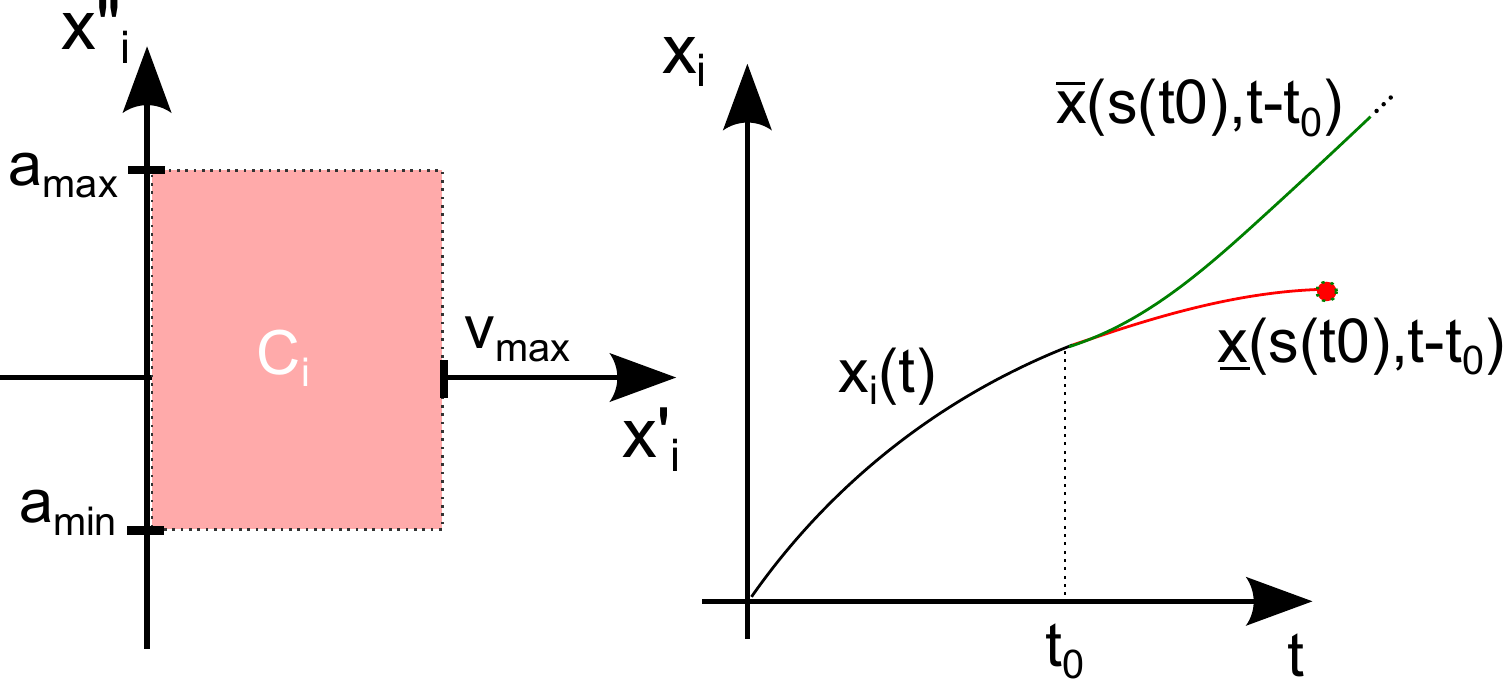}\hfill
\caption{The left drawing depicts an example of kinodynamic constraints where robots have uniform minimal/maximal velocity and acceleration along their paths. The right drawing depicts the corresponding minimal/maximal trajectories $\underline{x}(s(t_0),t)$ and $\overline{x}(s(t_0),t)$.}
\label{fig-constraints}
\end{figure}

The minimal (resp. maximal) trajectory from state $s^0$, denoted $\underline{x}(s_0,t)$ (resp. $\overline{x}(s_0,t)$), are defined bellow: 
\begin{eqnarray}
\underline{x}(s_0,t) & = & \min \chireach(s_0,t) \nonumber \\
\overline{x}(s_0,t) & = & \max \chireach(s_0,t) \nonumber
\end{eqnarray}
These are the lower and upper bounds of the hypercube $\chireach(s_0,t)$.
One can view the minimal trajectory as a braking trajectory, and the maximal trajectory as an accelerating trajectory. 
The concepts are illustrated by Figure \ref{fig-constraints} where the kinodynamic constraints have the special following form:
\begin{equation}
{C_i^\mathrm{acc}=\left\{ \left(x_i,x_i^\prime,x_i^{\prime\prime}\right) \left|
\begin{array}{l l}
0 \leq x_i' \leq v_i^\mathrm{max}\\
a_i^\mathrm{min} \leq x_i^{\prime\prime} \leq a_i^\mathrm{max} 
\end{array}
\right. \right\}}
\end{equation}

\subsection{The motion planner}
 
The time $t$ is first discretized, and the trajectory of the robots $x(t)$ is computed iteratively as described in the following Algorithm \ref{alg-fixed-priorities-with-dynamics}. Indeed, at every time-step $t$, the trajectory up to time $t+\Delta T$ can be computed as follows:
\begin{itemize}
\item Cycling through all robots, we select a particular robot $i$ (line \ref{line-selection-robot});
\item We compute a virtual path $\chi^\mathrm{virtual}$ (line \ref{line-chi-virtual}) that would be followed by the robots if:
\begin{itemize}
\item in the next time step, robot $i$ accelerates as much as possible while all other robots decelerate as much as possible (lines  6-11);
\item afterwards, all robots including $i$ brake as much as possible (see the second term of the concatenation at line \ref{line-chi-virtual})
\end{itemize}
\item  If this virtual path is such that no collision and priorities are respected, it means there exists a failsafe maneuver such that robot $i$ accelerates as much as possible in the next time step, and we let it do so. Otherwise, robot $i$ must brake (lines 14-18). Thus, at each time-step $t$, each robot $i$ exclusively follows its maximal or minimal trajectory  in the next time-step.
\end{itemize}

\begin{algorithm}[h]
\begin{algorithmic}[5]
\Require $s^\mathrm{init}, \text{ feasible priority graph } G$
\Function{maximallyAgressiveTrajectory}{}
\State $T \gets 0$
\State $s(0) \gets s^\mathrm{init}$
\While{$x(T)\notin \chigoal$}
\For{$i\in\otn$} \label{line-selection-robot}
\For{$t\in[0,\dt]$} \label{line-first-line-max-aggressive} 
\For{$j\neq i$}
\State $\tilde{s}_j(t) \gets {\underline{s}_j}(s(T),t)$
\EndFor
\State $\tilde{s}_i(t) \gets {\overline{s}_i}(s(T),t)$
\EndFor \label{line-last-line-max-aggressive}
\State $\chi^\mathrm{virtual} \gets \pi(\tilde{s}([0,\dt])) \cup  \underline{x}(\tilde{s}(\dt),\RR_+)$ \label{line-chi-virtual}
\If{$\exists (j,i)\in E(G)$ s.t. $\chi^\mathrm{virtual} \cap \chiobs_{j\succ i} \neq \emptyset$}
\State $s_i(T+\dt)\gets{\underline{s}_i}(s(T),\dt)$
\Else
\State $s_i(T+\dt)\gets{\overline{s}_i}(s(T),\dt)$
\EndIf
\label{line-end-planning-i}
\EndFor
\State $T\gets T+\Delta T$
\EndWhile
\State \Return{$(x(t))_{t=0\cdots T}$}
\EndFunction
\end{algorithmic}
\caption{The motion planner with assigned priorities}
\label{alg-fixed-priorities-with-dynamics}
\end{algorithm} 

The defined trajectory thus appears as a natural extension of the left-greedy trajectory introduced in \cite{Ghrist2006}, in the sense that in the absence of kinodynamic constraints ($p=1$), it coincides with it. Indeed, in this case the robots can stop instantly and the block from Line \ref{line-first-line-max-aggressive} to Line \ref{line-end-planning-i} simply consists of checking that maximum speed during the next time-step is safe: if it is not the case the robot is stopped. 

Note also that if the state $s(t)$ is such that the braking trajectory from $s(t)$ is collision-free, the state $s(t)$ is not an "Inevitable Collision State" (ICS), as defined in \cite{Fraichard2004} because the braking trajectory is collision-free, that is, there exists a particular collision-free trajectory starting from state $s(t)$. 

\subsection{Safety guarantees}

The theorem below exhibits the safety guarantee provided by the proposed motion planner.

\begin{theorem}[Safety guarantees]
Assume that there exists some feasible trajectory respecting priorities defined by $G$ and the initial state $s^\init$ is such that the initial braking trajectory $\underline{x}(s^\init,t)$ is collision-free. Then, for sufficiently small $\dt$, Algorithm \ref{alg-fixed-priorities-with-dynamics} terminates and returns a (collision-free) feasible trajectory respecting priorities $G$.
\end{theorem}

\begin{proof}
It is assumed that that there exists some feasible trajectory respecting priorities defined by $G$, so that $G$ is a feasible priority graph as defined in \cite{Gregoire2012}, and the trajectory cannot reach a deadlock configuration (for sufficiently small $\dt$). Until $\chigoal$ is reached, at any time there is at least one robot $i_0$ at a coordinate lower than $\sup\{x_{i_0}: x\in\chiobs\}$ moving forward. Indeed if this was not true it would mean that the robots  have reached a deadlock configuration. There is thus a lower bound, say $\mu$, for the distance travelled by some of the robots in a time-length $\dt$, depending on the constraints. This implies $x$ necessarily reaches $\chigoal$ in finite time (of order at most $O(n/\mu)$) and Algorithm \ref{alg-fixed-priorities-with-dynamics} terminates.

Now we prove that, at every time step, the braking trajectory from the current state is collision-free. We begin with a preliminary useful property, that is a direct consequence of  the definition of $\chiobs_{j \succ i}$ and is easily seen on Figure~\ref{fig-fixed-priorities-collision-cylinder}.
\begin{property}
Given $i,j\in\otn$ and two configurations $x,y\in\chi$ satisfying $y_j \geq x_j$ and $y_i \leq x_i$, we have:
\begin{equation}
x\in\chifree_{j\succ i} \Rightarrow y\in\chifree_{j\succ i}
\end{equation}
\label{property-fixed-collision-cylinder}
\end{property} 
The initial braking trajectory $\underline{x}(s^\init,t)$ is assumed to be collision-free. Now, assume that for some $t_0=k \dt$, $\underline{x}(s(t_0),t)$ is collision-free. In the next time step, for any priority $(j,i)\in E(G)$, there are two options for the robot with lower priority: 
\begin{itemize}
\item either $i$ brakes as much as possible. The fact that $\underline{x}(s(t_0),t)$ is collision-free with $\chiobs_{j \succ i}$ implies that $\underline{x}(s(t_0+\dt),t)$ is also collision-free by Property~\ref{property-fixed-collision-cylinder}, since we have for all $t\geq 0$:
\begin{eqnarray}
\underline{x}_i(s(t_0+\dt),t)&=&\underline{x}_i(s(t_0),t+\dt)\\
\underline{x}_j(s(t_0+\dt),t)&\geq&\underline{x}_j(s(t_0),t+\dt)
\end{eqnarray}

\item or $i$ accelerates as much as possible; in this case, the virtual path $\chi^\mathrm{virtual}$ is collision-free with respect to $\chiobs_{j\succ i}$. Then, consider the state $\tilde{s}(\dt)$ defined as:
\begin{eqnarray}
\tilde{s}_j(\dt)&=&\underline{s}_j(s(t_0),\dt)\\
\tilde{s}_i(\dt)&=&\overline{s}_i(s(t_0),\dt)
\end{eqnarray}
Since the virtual path $\chi^\mathrm{virtual}$ is collision-free with $\chiobs_{j\succ i}$, $\underline{x}(\tilde{s}(\dt),t)$ is also collision-free (see the second term in the concatenation at Line \ref{line-chi-virtual}). It implies that $\underline{x}(s(t_0+\dt),t)$ is also collision-free by Property~\ref{property-fixed-collision-cylinder}, since we have for all $t\geq 0$:
\begin{eqnarray}
\underline{x}_i(s(t_0+\dt),t)&=&\underline{x}_i(\tilde{s}(\dt),t)\\
\underline{x}_j(s(t_0+\dt),t)&\geq&\underline{x}_j(\tilde{s}(\dt),t)
\end{eqnarray}
\end{itemize}

Hence, at every time step, the braking trajectory is collision-free. Now, we prove that there is no collision between time steps. Again, at every time step $t_0=k \dt$, the are two options: 

\begin{itemize}
\item either $i$ brakes as much as possible. The fact that $\underline{x}(s(t_0),t)$ is collision-free with $\chiobs_{j \succ i}$ implies that for $t\in[t_0,t_0+\dt]$, $s(t)$ is also collision-free by Property~\ref{property-fixed-collision-cylinder}, since we have for all $t\in[t_0,t_0+\dt]$:
\begin{eqnarray}
x_i(t)&=&\underline{x}_i(s(t_0),t)\\
x_j(t)&\geq&\underline{x}_j(s(t_0),t)
\end{eqnarray}

\item or $i$ accelerates as much as possible; in this case, the virtual path $\chi^\mathrm{virtual}$ is collision-free with respect to $\chiobs_{j\succ i}$. Then, consider the trajectory $\tilde{x}(t)$ for $t\in[0,\dt]$ defined as:
\begin{eqnarray}
\tilde{x}_j(t)&=&\underline{x}_j(s(t_0),t)\\
\tilde{x}_i(t)&=&\overline{x}_i(s(t_0),t)
\end{eqnarray}
Since $\chi^\mathrm{virtual}$ is collision-free with $\chiobs_{j\succ i}$, $\tilde{x}(t)$ is also collision-free (see the first term in the concatenation at Line \ref{line-chi-virtual}). It implies that $s(t)$ is also collision-free for $t\in[t_0,t_0+\dt]$ by Property~\ref{property-fixed-collision-cylinder}, since we have:
\begin{eqnarray}
s_i(t)&=&\tilde{s}_i(t-t_0)\\
s_j(t)&\geq&\tilde{s}_j(t-t_0)
\end{eqnarray}
\end{itemize}

As a result, $x$ is collision-free with respect to $\chiobs_G$ at every time-step and reaches $\chigoal$: it is a collision-free feasible trajectory respecting priorities $G$.
\end{proof}

\section{Simulations}
\label{sec-simulations}

The algorithms presented in this paper have been implemented into a simulator coded in Java. Our algorithms have proved their ability to run in real-time. 

\subsection{Setting and results}
Only straight paths are implemented (for simplicity's sake) and all robots are supposed to be circle-shaped with a common radius $R$. The kinodynamic constraints of the robots concern only the maximal velocity and minimal/maximal acceleration. Moreover, all robots are supposed to have identical kinodynamic constraints. 
\begin{equation}
\forall i\in\otn, {C_i^\mathrm{acc}=\left\{ \left(x_i,x_i^\prime,x_i^{\prime\prime}\right) \left|
\begin{array}{l l}
0 \leq x_i' \leq v^\mathrm{max}\\
a^\mathrm{min} \leq x_i^{\prime\prime} \leq a^\mathrm{max} 
\end{array}
\right. \right\}}\text{.}
\end{equation}

In the simulation results presented in this section, we take as priority assignment policy the maximally aggressive priority assignment policy that consists for every robot of taking priority over another robot if it reaches the conflicting region first. This priority assignment policy can lead to deadlock configurations (see \cite{Gregoire2012}), but with a very small probability in case of low traffic density as in the presented simulations. This policy is used for the sake of simplicity, the priority assignment policy not being the focus of this paper. 

\begin{figure}[ht]
\centering
\includegraphics[width=1\linewidth]{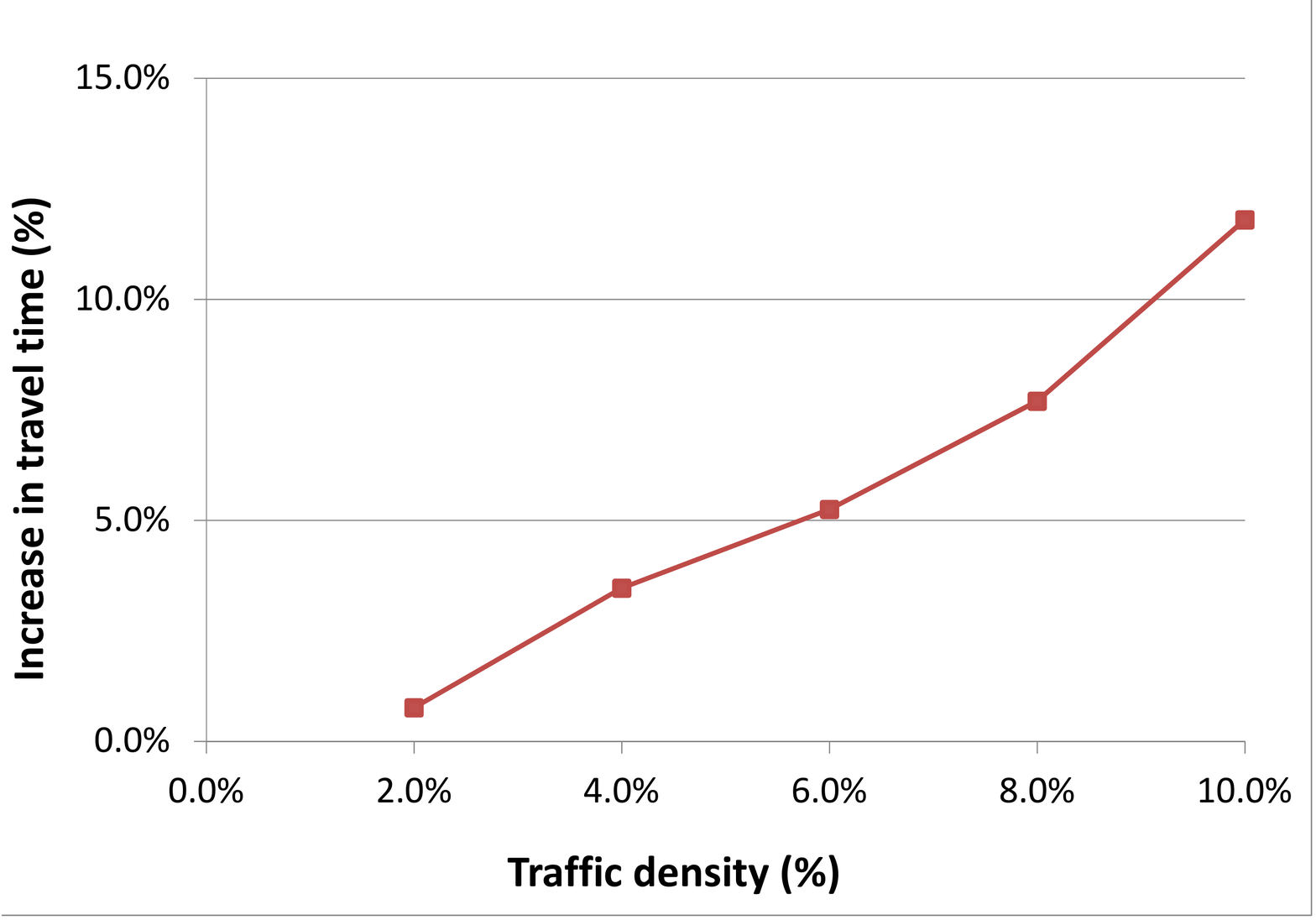}\hfill
\caption{Simulation results: plot of the averaged increase in travel time against the traffic density for the intersection of Figure \ref{fig-intersection-simulation}}
\label{fig-performance-results}
\end{figure}

\begin{figure}[ht]
\centering
\includegraphics[width=0.5\linewidth]{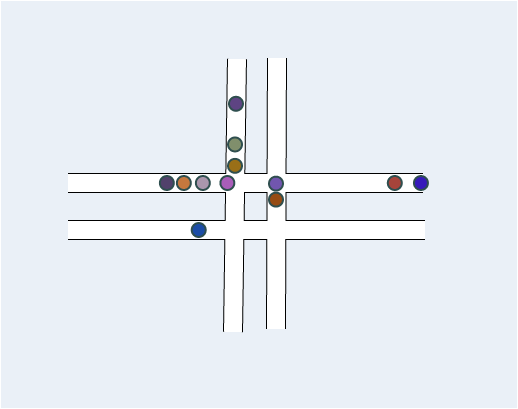}\hfill\\
\caption{The 4-path-intersection used for simulations}
\label{fig-intersection-simulation}
\end{figure}

Simulations have been carried out for the 4-path-intersection depicted in Figure \ref{fig-intersection-simulation}. At full speed, the distance travelled in one time-step is $R$ and at full acceleration, 20 time-steps are required for the robots to reach full speed. Figure \ref{fig-performance-results} depicts the increase in travel time for different traffic densities. The increase in travel time is the delay due to coordination, i.e. the difference with the ideal travel time which is the travel time of robots in the absence of other robots. It is expressed in percentage of the ideal travel time. The increase in travel time vanishes as the density approaches 0 since it becomes very unlikely that they need to coordinate to avoid collisions. The traffic density in percentage is the ratio between the actual traffic density and the maximum traffic density (continuous flow of robots). The robots are generated randomly at a constant rate over time. The video of the simulation for a traffic density of 10\% is available at \url{http://youtu.be/bJHdf3AbIlI}.

\subsection{Comments}

First of all, our algorithm succeeds to work in real time, and one can observe in simulations (notably on the video) that collisions never occur.  This confirms the fact that the planner guarantees safety under dynamic constraints. One can see in Figure \ref{fig-performance-results} that at a traffic density of $10\%$ on each path, the increase in travel time due to coordination to avoid other robots is less than $15\%$ which seems a low price to pay to ensure safe coordination. Note that we do not present simulation results at higher traffic densities because it would require to define a more complex priority assignment policy (at least to avoid deadlocks), which is a challenge in itself, and beyond the scope of the present paper. 

\section{Conclusions and discussion}
\label{sec-conclusion}

The results presented in this paper prove that when priorities are assigned, it is possible to plan a safe and quite efficient trajectory respecting the priority graph and the dynamic constraints of the robots. The use of the braking trajectory enables to anticipate the need to brake just-in-time, and as a byproduct provides robustness guarantees since there exists a collision-free braking maneuver at any time.

If the robots drift from the planned trajectory but if no priority has been violated, it is possible to run the motion planner from a new initial state to get a new feasible trajectory respecting the assigned priorities. This reflects that the method proposed in this paper is inherently a feedback motion planning approach (see \cite{LaValle2006}, chapter 8). We are currently turning the planning algorithms of this paper into a feedback control law that aims at coordinating robots with assigned priorities. The idea is to define a control law $g^G(s)$ that maps every state $s$ to the control to apply in the next time step. The control law $g^G$ is in charge of coordination, ensuring that collisions are avoided and that priorities $G$ are respected. Robots do not have to follow precisely a planned trajectory, they just have to be aware of the priorities and to respect the control law. The benefit of the approach is that it ensures safe coordination  as long as priorities $G$ are respected, which is much easier to robustly ensure than following precisely a planned trajectory. This opens avenues to build multiple robot coordination systems much more robust with regards to uncertainty in control and sensing.

\bibliographystyle{IEEEtran}
\bibliography{biblio}

\begin{thebibliography}{10}
\providecommand{\url}[1]{#1}
\csname url@samestyle\endcsname
\providecommand{\newblock}{\relax}
\providecommand{\bibinfo}[2]{#2}
\providecommand{\BIBentrySTDinterwordspacing}{\spaceskip=0pt\relax}
\providecommand{\BIBentryALTinterwordstretchfactor}{4}
\providecommand{\BIBentryALTinterwordspacing}{\spaceskip=\fontdimen2\font plus
\BIBentryALTinterwordstretchfactor\fontdimen3\font minus
  \fontdimen4\font\relax}
\providecommand{\BIBforeignlanguage}[2]{{%
\expandafter\ifx\csname l@#1\endcsname\relax
\typeout{** WARNING: IEEEtran.bst: No hyphenation pattern has been}%
\typeout{** loaded for the language `#1'. Using the pattern for}%
\typeout{** the default language instead.}%
\else
\language=\csname l@#1\endcsname
\fi
#2}}
\providecommand{\BIBdecl}{\relax}
\BIBdecl

\bibitem{Treat1977}
J.~Treat, N.~Castellan, R.~Stansifer, R.~Mayer, R.~Hume, D.~Shinar,
  S.~McDonald, and N.~Tumbas, \emph{Tri-level Study of the Causes of Traffic
  Accidents: Final Report. Volume I: Causal Factor Tabulations and
  Assessments}, 1977.

\bibitem{NCSA2004}
NCSA, ``National center for statistics and analysis, traffic safety facts
  2003,'' U.S. DOT, Washington, DC, Tech. Rep., 2004.

\bibitem{Dresner2004}
K.~Dresner and P.~Stone, ``Multiagent traffic management: A reservation-based
  intersection control mechanism,'' in \emph{Proceedings of the Third
  International Joint Conference on Autonomous Agents and Multiagent
  Systems-Volume 2}, july 2004, pp. 530 --537.

\bibitem{Dresner2008}
------, ``Mitigating catastrophic failure at intersections of autonomous
  vehicles,'' in \emph{{AAMAS} Workshop on Agents in Traffic and
  Transportation}, Estoril, Portugal, May 2008, pp. 78--85.

\bibitem{Colombo2012}
A.~Colombo and D.~D. Vecchio, ``Efficient algorithms for collision avoidance at
  intersections,'' \emph{Hybrid Systems: Computation and Control}, 2012.

\bibitem{I.Zohdy2012}
I.~Zohdy and H.~Rakha, ``Optimizing driverless vehicles at intersections,'' in
  \emph{10th ITS World Congress Vienna, Austria}, October 2012.

\bibitem{Kowshik2011}
H.~Kowshik, D.~Caveney, and P.~Kumar, ``Provable systemwide safety in
  intelligent intersections,'' \emph{IEEE Transactions on Vehicular
  Technology}, vol.~60, no.~3, pp. 804 --818, march 2011.

\bibitem{Mehani2007}
O.~Mehani and A.~De~La~Fortelle, ``Trajectory planning in a crossroads for a
  fleet of driverless vehicles,'' in \emph{Proceedings of the 11th
  international conference on Computer aided systems theory}, ser.
  EUROCAST'07.\hskip 1em plus 0.5em minus 0.4em\relax Berlin, Heidelberg:
  Springer-Verlag, 2007, pp. 1159--1166.

\bibitem{Ismail1999}
Ismail and Sahin, ``Railway traffic control and train scheduling based
  oninter-train conflict management,'' \emph{Transportation Research Part B:
  Methodological}, vol.~33, no.~7, pp. 511 -- 534, 1999.

\bibitem{Tomlin1998}
C.~Tomlin, G.~Pappas, and S.~Sastry, ``Conflict resolution for air traffic
  management: a study in multiagent hybrid systems,'' \emph{IEEE Transactions
  on Automatic Control}, vol.~43, no.~4, pp. 509 --521, apr 1998.

\bibitem{Kant1986}
K.~Kant and S.~W. Zucker, ``Toward efficient trajectory planning: The
  path-velocity decomposition,'' \emph{International Journal of Robotics
  Research}, vol.~5, no.~3, pp. 72--89, 1986.

\bibitem{Gregoire2012}
J.~Gregoire, S.~Bonnabel, and A.~De~La~Fortelle, ``Optimal cooperative motion
  planning for vehicles at intersections,'' in \emph{Navigation, Perception,
  Accurate Positioning and Mapping for Intelligent Vehicles, Workshop, 2012
  IEEE Intelligent Vehicles Symposium}, 2012.

\bibitem{Leroy1999}
S.~Leroy, J.~P. Laumond, and T.~Simeon, ``Multiple path coordination for mobile
  robots: A geometric algorithm,'' in \emph{Proceedings of the International
  Joint Conference on Artificial Intelligence (IJCAI)}, 1999, pp. 1118--1123.

\bibitem{Lavalle1996}
S.~LaValle and S.~Hutchinson, ``Optimal motion planning for multiple robots
  having independent goals,'' in \emph{Proceedings of the IEEE International
  Conference on Robotics and Automation, 1996}, vol.~3, apr 1996, pp. 2847
  --2852 vol.3.

\bibitem{Latombe1991}
J.-C. Latombe, \emph{Robot Motion Planning}.\hskip 1em plus 0.5em minus
  0.4em\relax Norwell, MA, USA: Kluwer Academic Publishers, 1991.

\bibitem{Lozano-Perez1980}
T.~Lozano-Perez, ``Spatial planning: A configuration space approach,'' 1980.

\bibitem{LaValle2006}
S.~M. LaValle, \emph{Planning Algorithms}.\hskip 1em plus 0.5em minus
  0.4em\relax Cambridge, U.K.: Cambridge University Press, 2006, available at
  http://planning.cs.uiuc.edu/.

\bibitem{Ghrist2006}
R.~Ghrist and S.~M. Lavalle, ``Nonpositive curvature and pareto optimal
  coordination of robots,'' \emph{SIAM Journal on Control and Optimization},
  vol.~45, pp. 1697--1713, November 2006.

\bibitem{Peng2005}
J.~Peng and S.~Akella, ``Coordinating multiple robots with kinodynamic
  constraints along specified paths,'' \emph{The International Journal of
  Robotics Research}, vol.~24, no.~4, pp. 295--310, 2005.

\bibitem{Fraichard1989}
T.~Fraichard and C.~Laugier, ``Planning movements for several coordinated
  vehicles,'' in \emph{IEEE/RSJ International Workshop on Intelligent Robots
  and Systems '89. The Autonomous Mobile Robots and Its Applications. IROS
  '89.}, sep 1989, pp. 466 --472.

\bibitem{Hafner2011}
C.~Hafner, Cunningham and D.~Vechhio, ``Automated vehicle-to-vehicle collision
  avoidance at intersections,'' 2011.

\bibitem{Akella2002}
S.~Akella and S.~Hutchinson, ``Coordinating the motions of multiple robots with
  specified trajectories,'' in \emph{Proceedings of the IEEE International
  Conference on Robotics and Automation, 2002. ICRA '02.}, vol.~1, 2002, pp.
  624 -- 631 vol.1.

\bibitem{ODonnell1989}
P.~O'Donnell and T.~Lozano-Periz, ``Deadlock-free and collision-free
  coordination of two robot manipulators,'' in \emph{Proceedings of the IEEE
  International Conference on Robotics and Automation, 1989.}, may 1989, pp.
  484 --489 vol.1.

\bibitem{Fraichard2004}
T.~Fraichard and H.~Asama, ``\BIBforeignlanguage{Anglais}{{Inevitable collision
  states - a step towards safer robots?}}''
  \emph{\BIBforeignlanguage{Anglais}{Advanced Robotics -Utrecht-}}, vol.~18,
  no.~10, pp. 1001--1024, 2004.

\end{thebibliography}

\end{document}